\documentclass[11pt]{article}

\usepackage[final]{acl}

\usepackage{times}
\usepackage{latexsym}
\usepackage{amssymb}
\usepackage{pifont}
\usepackage{multirow} 
\usepackage{enumitem}

\usepackage{CJKutf8}
\usepackage{siunitx}
\usepackage[table]{xcolor}
\usepackage[T1]{fontenc}
\usepackage{booktabs}   
\usepackage{xcolor}     
\usepackage{array}      

\usepackage[utf8]{inputenc}

\usepackage{microtype}
\usepackage{amsmath}
\usepackage{inconsolata}

\usepackage{graphicx}

\title{
SONAR-SLT: Multilingual Sign Language Translation via Language-Agnostic Sentence Embedding Supervision
}

\author{
\textbf{Yasser Hamidullah}$^{1}$\quad
\textbf{Shakib Yazdani}$^{1}$\quad
\textbf{Cennet Oguz}$^{1}$\\
\textbf{Josef van Genabith}$^{1}$\quad
\textbf{Cristina España-Bonet}$^{1,2}$\\[6pt]
$^{1}$German Research Center for Artificial Intelligence (DFKI GmbH),\\
Saarland Informatics Campus, Saarbrücken, Germany \\
$^{2}$Barcelona Supercomputing Center (BSC-CNS), Barcelona, Catalonia, Spain\\[6pt]
\texttt{\{yasser.hamidullah,shakib.yazdani,cennet.oguz,josef.van\_genabith,cristinae\}@dfki.de}
}

\begin{document}
\maketitle

\begin{abstract}
Sign language translation (SLT) is typically trained with text in a single spoken language, which limits scalability and cross-language generalization. Earlier approaches have replaced gloss supervision with text-based sentence embeddings, but up to now, these remain tied to a specific language and modality. In contrast, here we employ language-agnostic, multimodal embeddings trained on text and speech from multiple languages to supervise SLT, enabling direct multilingual translation. 
To address data scarcity, we propose a coupled augmentation method that combines multilingual target augmentations (i.e. translations into many languages) with video-level perturbations, improving model robustness. Experiments show consistent BLEURT gains over text-only sentence embedding supervision, with larger improvements in low-resource settings. Our results demonstrate that language-agnostic embedding supervision, combined with coupled augmentation, provides a scalable and semantically robust alternative to traditional SLT training.%
\footnote{We release the code, models, and features to facilitate further research. Github repository: \href{https://github.com/DFKI-SignLanguage/sonar-slt.git}{https://github.com/DFKI-SignLanguage/sonar-slt.git}; Huggingface: \href{https://huggingface.co/mtmlt}{https://huggingface.co/mtmlt}}

\end{abstract}

\section{Introduction}
Sign languages (SLs) are inherently visual and culturally embedded. Each SL has evolved independently and is closely tied to the communities and spoken languages of its region. As a result, most sign language translation (SLT) datasets are built around a \emph{single} sign--spoken language pair (e.g., DGS$\rightarrow$German), which makes it difficult to scale models across languages or to combine datasets. Training a system for a new target language typically requires a separate model and fresh parallel data collection.

Historically, SLT systems have relied on manually provided \emph{gloss supervision} \citep{Camgoz_2018_CVPR}, discrete word-like labels whose design and availability are language-, culture-, and region-specific. Even \textit{gloss-free} SLT approaches assume that sign inputs should be supervised by text from the co-occurring spoken language, keeping the learning signal tied to a single language \citep{gong2024signllm, wong2024sign2gpt, chen-etal-2024-factorized, hamidullah2022spatio} and limiting cross-dataset reuse and generalization.

\begin{figure}[t]
  \includegraphics[width=\columnwidth]{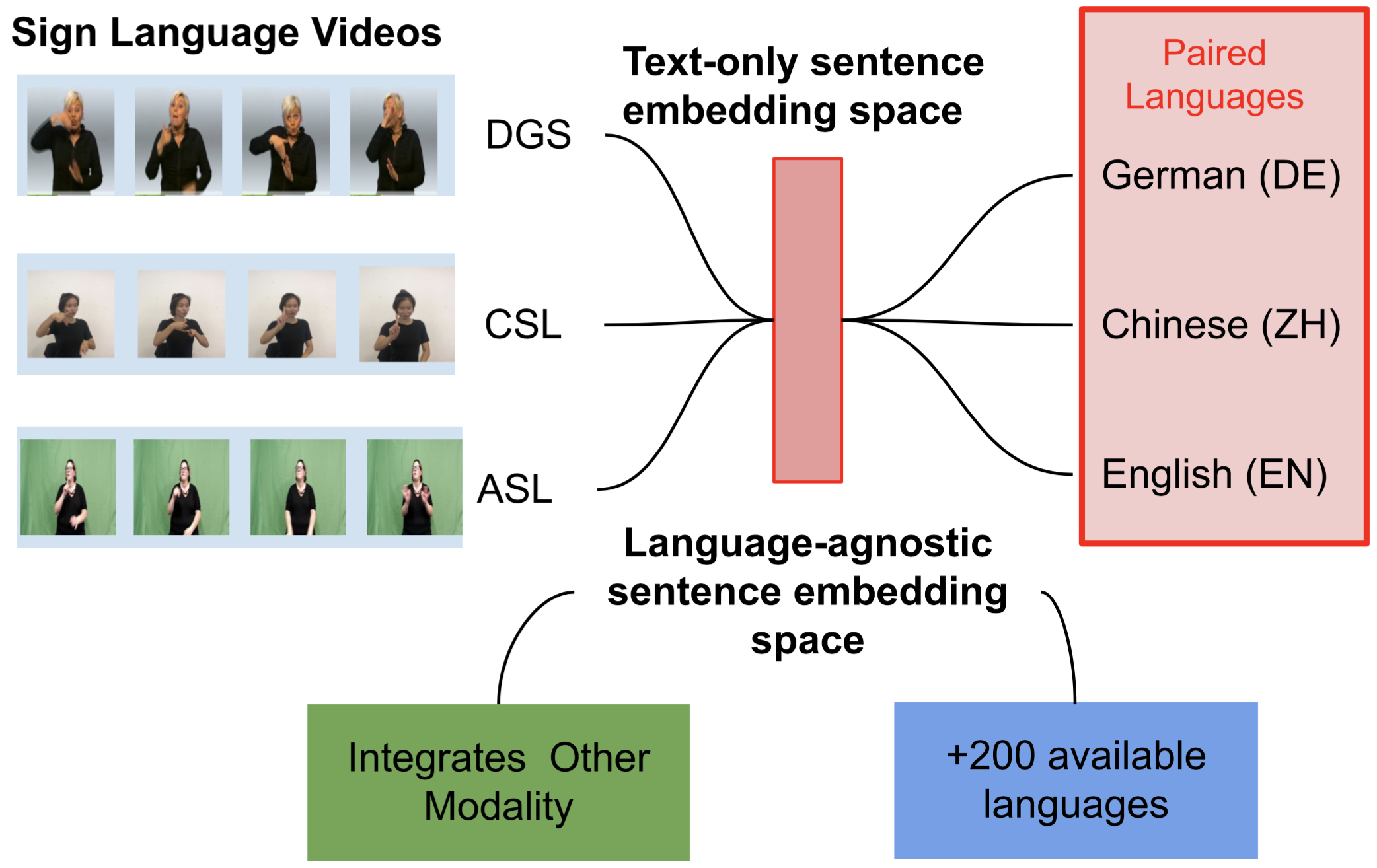}
  \caption{Text-only vs. language-agnostic sentence embedding supervision.}
    \label{fig:mini_diag}
\end{figure}

Recent work by \citet{hamidullah2024slt} has reduced the reliance on glosses by supervising SLT with \emph{text-based sentence embeddings}. This yields better semantic alignment, but the embeddings remain modality-specific and typically require dataset-specific fine-tuning. Furthermore, compared to large pre-trained models that exploit vast text corpora, these text-only embeddings show limited cross-lingual transfer and reduced robustness. This raises the key question: \emph{Can language-agnostic, multimodal sentence embedding supervision replace text-only alignment in SLT?} We hypothesize that \textbf{language-agnostic, multimodal sentence embeddings} can reduce the residual dependence on text. Concretely, we build on \textsc{SONAR} \citep{duquenne2023sonar}, a pretrained multilingual and multimodal embedding space that jointly represents text and speech. SONAR embeddings are claimed to be language-agnostic.
Our approach aligns sign representations directly with language-agnostic semantic vectors, thereby decoupling supervision from any specific spoken language and removing the need for glosses. Our model integrates multiple modalities and supports direct supervision across all 200 languages covered by \textsc{SONAR} (see Figure~\ref{fig:mini_diag}). In contrast to prior systems that relied on additional stages or separate models for multi-target translation, our method enables \emph{direct} translation into multiple languages within a single model.

A major obstacle for SLT is the scarcity of annotated data. Recent work on self-supervised pre-training from unannotated or anonymized data \citep{rust2024ssvp} has shown promise in addressing this challenge. This motivates our second question: \emph{Can target-language augmentation further alleviate data scarcity and enhance robustness, particularly when combined with video augmentation?}

Our \textbf{coupled multiple target language and video perturbation augmentation} strategy addresses these challenges by combining (i) \emph{target-language augmentation}, which pairs each sign sample with parallel sentences in multiple languages, and (ii) \emph{video augmentation}, which perturbs the visual stream through spatial, temporal, and photometric transformations. These augmentations are complementary: multiple target-language augmentation strengthens semantic supervision without requiring new sign recordings, while video augmentation improves the invariance of the sign encoder. Together, they yield a more robust SLT model and provide a scalable, semantically grounded alternative to traditional training, unifying supervision across languages and modalities while reducing dependence on language-, culture-, and region-specific annotations. In all, our contributions can be summarized as:

\begin{itemize}
\item \textbf{Language-agnostic supervision.} We align signs to a multilingual, multimodal embedding space, removing reliance on language-specific text or glosses.
\item \textbf{Coupled augmentation.} We jointly apply multilingual target augmentation and video perturbations to improve robustness and reduce data scarcity.
\item \textbf{Direct multilingual decoding.} Our model translates into multiple spoken languages in a single step, without pivots or extra fine-tuning.
\item \textbf{Open-source resources.} We release a Hugging Face–compatible \emph{visual extension of \textsc{SONAR}} and model port to enable reproducibility and further work.
\end{itemize}

\begin{figure*}[t]
    \centering
    \includegraphics[width=2.1\columnwidth]{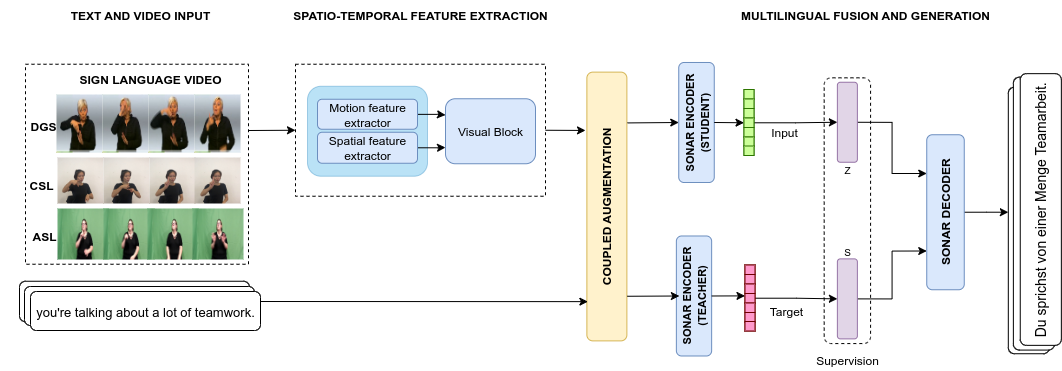}
  \caption{Overall architecture of our SONAR-SLT model. Visual inputs are processed through spatial and spatio-
temporal encoders, fused using \citep{hwang-etal-2025-spamo} and encoded into a semantic vector aligned with multilingual sentence embeddings.}
    \label{fig:architecture}
\end{figure*}

\section{Related Work}

\subsection{Sign Language Representation}

Traditional SLT systems rely on glosses ---textual labels that represent signs--- as an intermediate representation. MSKA-SLT~\citep{10.1016/j.patcog.2025.111602} remains a strong baseline using glosses, reporting $\sim$29 BLEU on PHOENIX-2014T \citep{Camgoz_2018_CVPR}. However, glosses are neither universal nor standardized: they are tightly coupled to specific languages, cultures, and regions. Moreover, producing gloss annotations is highly time-consuming, requiring expert linguistic knowledge \citep{muller-etal-2023-considerations}.

In parallel, gloss-free SLT has emerged, enabling training on weakly annotated datasets exceeding 1{,}000 hours for some sign languages \citep{uthus2023youtubeasl}.%
\footnote{A \emph{weakly annotated dataset} provides only coarse or noisy supervision. For instance, YouTube-ASL datasets are collected from online videos where annotations rely solely on automatically generated or the provided subtitles, without manual realignment, 
leading to potential inaccuracies and temporal misalignments.} 
\citet{hamidullah2024slt} aligns sign language videos with sentence-level text embeddings. This supervision avoids feeding long, fine-grained frame sequences to the decoder, thereby reducing redundancy in video features, lowering the need for aggressive masking, and encouraging learning at the sentence-semantic level. While intermediate supervision of visual blocks is common in multimodal models, compressing video into a sentence-level embedding before decoding improves semantic grounding and flexibility in target text generation. Nevertheless, current approaches \citep{hamidullah2024slt,shubert} remain limited by their reliance on \emph{text-only} embedding spaces with restricted language coverage, constraining augmentation and cross-sign transfer. 



\subsection{Large Language Models in SLT}
Complementary approaches leverage large language models (LLMs). SignLLM~\citep{gong2024signllm} discretizes videos into tokens and prompts a frozen LLM; Sign2GPT~\citep{wong2024sign2gpt} feeds pseudo-glosses to XGLM, reporting $\sim$22 BLEU on PHOENIX-2014T and $\sim$15 BLEU on CSL-Daily. SpaMo~\citep{hwang-etal-2025-spamo} employs a straightforward approach that extracts spatial and motion features from sign language videos and utilizes a low-rank adapter to fine-tune an LLM for sign language translation. \citet{chen-etal-2024-factorized} introduced FLa-LLM, a two-stage, gloss-free framework that first pre-trains the visual encoder and then fine-tunes a pre-trained LLM for the downstream SLT task. These methods inherit LLM fluency but are largely monolingual and require substantial tokenization and training overhead. In contrast, our PEFT-based SONAR adapters maintain multilinguality without retraining a large decoder on discretized video tokens. More recent work has explored large-scale pre-training to improve sign language understanding, with Uni-Sign \citep{li2025unisignunifiedsignlanguage} proposing a unified generative framework that treats downstream tasks as SLT and incorporates prior-guided fusion.

\subsection{Multilingual SLT Datasets and Models}
Despite these advances, large-scale multilingual datasets \citep{uthus2023youtubeasl, yazdani-EtAl:2025:RANLP} remain scarce and noisy. Crawled web data increases coverage but introduces label and alignment errors that current models struggle to absorb, leading many studies to focus on a single language or a small set of cleaner corpora. Additionally, performance often varies widely even within the same language due to differences in feature pipelines and recording conditions. 

Multilingual SLT models also remain in their early stages. MLSLT~\citep{yin2022multislt} covers ten European sign languages via a routing mechanism, while JWSign~\citep{gueuwou2023jwsign} scales to 98 languages with language-ID tokens. 
More recently, Sign2(LID+Text)~\citep{tan-etal-2025-multilingual} incorporated token-level language identification with a CTC loss, achieving competitive results. In addition, \citet{yazdani-etal-2025-continual} explored continual learning for multilingual SLT.
Recent work applies heavy pre-processing \citep{shubert, gueuwou-etal-2025-signmusketeers}, sometimes obscuring whether improvements arise from better SLT modeling or dataset-specific engineering. Both gloss-based and gloss-free methods perform best when signer distance, camera setup, and motion characteristics closely match training conditions.

\section{Methodology}
\label{sec:method}

\subsection{System Overview}
We propose \textbf{SONAR-SLT}, a modular SLT framework that decouples 
\emph{semantic understanding} from \emph{text generation}. 
As illustrated in Figure~\ref{fig:architecture}, the system first maps an input sign language video into a multilingual, multimodal semantic space, and then (optionally) decodes from this space into a chosen spoken language.
This design allows training on heterogeneous sign language datasets, supports multilingual supervision, and removes the need for gloss annotations. A detailed architecture is presented in Appendix \ref{app:arch} and summarized in the next subsections.

\subsection{Visual Feature Extraction and Encoding}
The first stage maps raw video frames into a compact visual embedding.  
Let $x=(f_1,\dots,f_T)$ denote a sign language video of $T$ frames. 
We extract per-frame spatial features $\mathbf{s}_t$ with ViT \citep{vitdosovitskiy2020image} and spatio-temporal 
motion features $\mathbf{m}_t$ with VideoMAE \citep{tong2022videomae}. These are fused through a lightweight block (1D Conv followed by a multi-layer perceptron)   
$\mathcal{F}$~\citep{hwang-etal-2025-spamo}:  
\begin{equation}
\mathbf{h}_t \;=\; \mathcal{F}\!\left(\mathbf{s}_t,\mathbf{m}_t\right),\qquad t=1,\dots,T.
\end{equation}
A Transformer-based encoder $\mathcal{E}_v$ contextualizes the sequence:
\begin{equation}
\mathbf{z}_{1:T} \;=\; \mathcal{E}_v(\mathbf{h}_{1:T}).
\end{equation}
Finally, temporal pooling (mean or attention) produces a global 
visual embedding $\mathbf{z}\in\mathbb{R}^d$:  
\begin{equation}
\mathbf{z} \;=\; \mathrm{Pool}(\mathbf{z}_{1:T}).
\end{equation}

\subsection{Semantic Alignment}
\label{sec:sub_sem_align}
Next, we align sign-derived embeddings with multilingual textual embeddings.  
We adopt a pretrained multilingual, multimodal sentence encoder $\mathcal{E}$ 
(i.e., \textsc{SONAR}). 
Given a reference sentence $y$, we obtain its semantic embedding:  
\begin{equation}
\mathbf{s} \;=\; \mathcal{E}_{txt}(y)\in\mathbb{R}^d.
\end{equation}
The visual encoder is trained to align $\mathbf{z}$ with $\mathbf{s}$. Alignment can be done via 
a squared $\ell_2$ loss as per \citep{duquenne2023sonar, hamidullah2024slt}:  
\begin{equation}
\label{eq:sem}
\mathcal{L}_{\text{sem}} \;=\; \big\|\mathbf{z}-\mathbf{s}\big\|_2^2.
\end{equation}

We also consider a cosine similarity loss,  
\begin{equation}
\label{eq:cos}
\mathcal{L}_{\cos} \;=\; 1 \;-\; \frac{\langle \mathbf{z}, \mathbf{s}\rangle}{\|\mathbf{z}\|_2\,\|\mathbf{s}\|_2},
\end{equation}
used either alone ($\mathcal{L}_{\text{sem}}=\mathcal{L}_{\cos}$) or 
combined with the MSE above:  
\begin{equation}
\label{eq:comb}
\mathcal{L}_{\text{sem}} \;=\; \alpha\,\|\mathbf{z}-\mathbf{s}\|_2^2 \;+\; \beta\,\mathcal{L}_{\cos}, 
\qquad \alpha,\beta \ge 0.
\end{equation}

\paragraph{Target-language augmentation.}
To enforce language-agnostic supervision, each reference sentence is paired with $K$ 
translations $\{y^{(k)}\}_{k=1}^{K}$ (from the embedding decoder). At each iteration, one translation $y^{(k)}$ is sampled 
and encoded as $\mathbf{s}=\mathcal{E}_{txt}(y^{(k)})$.  

\subsection{Multilingual Generation from the Semantic Vector}
We then decode into natural language from the semantic embedding.  
A pretrained decoder $\mathcal{D}$ from \textsc{SONAR} generates text from a semantic vector 
and a target language token $\ell$. Conditioned on the sign-derived and semantically text-aligned (Section ~\ref{sec:sub_sem_align}) embedding $\mathbf{z}$, 
the decoder is trained with teacher forcing:  
\begin{equation}
\label{eq:ce}
\mathcal{L}_{\text{ce}} \;=\; -\sum_{t=1}^{T_y} \log p_\theta\!\left(y_t \mid y_{<t},\,\mathbf{z},\,\ell\right).
\end{equation}

\subsection{Auto-Encoding (Decoder Anchoring)}
To keep the decoder aligned to the pretrained semantic space, we introduce an auto-encoding step.  
Specifically, the decoder reconstructs the target sentence directly from its text-derived embedding $\mathbf{s}$:  
\begin{equation}
\label{eq:ae}
\mathcal{L}_{\text{ae}} \;=\; -\sum_{t=1}^{T_y} \log p_\theta\!\left(y_t \mid y_{<t},\,\mathbf{s},\,\ell\right).
\end{equation}
This mirrors \textsc{SONAR}'s original training and prevents drift, while the visual encoder learns to project videos into the same space.  

\subsection{Optional Contrastive Alignment}
We optionally strengthen alignment through a symmetric InfoNCE loss~\citep{Oord2018RepresentationLW}. 
For a batch $\{(\mathbf{z}_i,\mathbf{s}_i)\}_{i=1}^{N}$, we define similarity as  
$\mathrm{sim}(\mathbf{a},\mathbf{b})=\frac{\mathbf{a}^\top \mathbf{b}}{\tau}$ 
(temperature $\tau>0$, optional $\ell_2$ normalization). The corresponding loss is:  
\begin{equation}
\label{eq:nce-broken}
\begin{aligned}
\mathcal{L}_{\text{nce}}
&= \frac{1}{2N}\sum_{i=1}^{N}\Bigg[
-\log \frac{\exp(\hat{\mathbf{z}}_i^\top \hat{\mathbf{s}}_i/\tau)}
{\sum_{j=1}^{N}\exp(\hat{\mathbf{z}}_i^\top \hat{\mathbf{s}}_j/\tau)} \\[-2pt]
&\hphantom{= \frac{1}{2N}\sum_{i=1}^{N}\Bigg[}
-\log \frac{\exp(\hat{\mathbf{s}}_i^\top \hat{\mathbf{z}}_i/\tau)}
{\sum_{j=1}^{N}\exp(\hat{\mathbf{s}}_i^\top \hat{\mathbf{z}}_j/\tau)}
\Bigg].
\end{aligned}
\end{equation}

\subsection{Joint Training Objective}
The final loss combines all components:  
\begin{align}
\label{eq:joint}
\mathcal{L}_{\text{joint}}
\;=\; &
\lambda_{\text{sem}}\,\mathcal{L}_{\text{sem}}
+\lambda_{\text{ce}}\,\mathcal{L}_{\text{ce}} \nonumber\\
&+\lambda_{\text{ae}}\,\mathcal{L}_{\text{ae}}
+\lambda_{\text{nce}}\,\mathcal{L}_{\text{nce}},
\end{align}
with non-negative weights $\lambda_i$ 
(setting $\lambda_{\text{nce}}=0$ disables the contrastive term).

\begin{table*}[h]
\centering
\footnotesize
\begin{tabular}{lcccccc}
\toprule
Dataset & Language & Domain & \#Videos & \#Sent. & Vocab. & Split (train/dev/test) \\
\midrule
PHOENIX-2014T & DGS $\rightarrow$ German & Weather Forecast & $\sim$7k & $\sim$8k & $\sim$3k & 7,096 / 519 / 642 \\
CSL-Daily     & CSL $\rightarrow$ Chinese & Daily Communication & $\sim$20k & $\sim$25k & $\sim$5k & 18,401 / 1,078 / 1,057 \\
\bottomrule
\end{tabular}
\caption{Characteristics of the datasets used in our experiments.}
\label{tab:datasets}
\end{table*}

\subsection{Cross-Lingual and Multi-Sign Dataset Fusion}
Finally, we leverage the language-agnostic semantic space for dataset fusion.  
Because supervision is defined independently of any specific spoken language, videos from different sign languages can be trained jointly with textual supervision in \emph{any} available language. For example, German sign language videos annotated in German can 
be re-aligned with English, French, or Chinese translations via $\mathcal{E}$, allowing unified training across datasets such as PHOENIX-2014T and CSL-Daily.  
This enables direct multi-target translation without glosses and facilitates fusion of heterogeneous sign-language corpora.  

\section{Experiments}
\subsection{Datasets}
We evaluate our approach on the following datasets:
\begin{itemize}
\item \textbf{PHOENIX-2014T} \citep{Camgoz_2018_CVPR}: German Sign Language (DGS) weather forecast videos with parallel German text. 

\item \textbf{CSL-Daily} \citep{Zhou_2021_CVPR}: A Chinese Sign Language (CSL) corpus tailored for sign-to-Chinese SLT, emphasizing interactions in daily communication contexts.
\end{itemize}

Statistics of both datasets are summarizes in Table~\ref{tab:datasets}.

\subsection{Evaluation Metrics}
We evaluate our method following \citep{muller-etal-2022-findings,muller-etal-2023-findings}, 
using BLEU\footnote{\texttt{BLEU|nrefs:1|bs:1000|seed:16|case: mixed|eff:no|tok:13a|smooth:exp|version:2.4.0}} 
(via SacreBLEU \citep{post-2018-call}) for lexical overlap, 
ROUGE \citep{lin-2004-rouge}\footnote{\texttt{ROUGE|L|nrefs:1|tok:13a|case:mixed|version:1.5.5}} 
for recall-oriented $n$-gram overlap, 
and BLEURT \citep{sellam-etal-2020-bleurt}\footnote{BLEURT v0.0.2 using checkpoint BLEURT-20.} 
for semantic quality.

\subsection{State-of-the-art Systems}
We evaluate our method against several strong recent state-of-the-art systems within the gloss-free paradigm. CSGCR~\citep{zhao2021conditional} improves SLT accuracy and fluency through three modules: word existence verification, conditional sentence generation, and cross-modal re-ranking for richer grammatical representations. GFSLT-VLP~\citep{Zhou_2023_ICCV} leverages vision–language pretraining, while FLa-LLM~\citep{chen-etal-2024-factorized} adopts a two-stage gloss-free pipeline that first pre-trains the visual encoder and then fine-tunes a pre-trained LLM for SLT. Sign2GPT~\citep{wong2024sign2gpt} maps visual inputs to pseudo-gloss sequences and decodes them with GPT-style language modeling, whereas SignLLM~\citep{gong2024signllm} discretizes sign features into visual tokens to prompt a frozen LLM. SEM-SLT~\citep{hamidullah2024slt} aligns sign language videos with sentence embeddings and serves as the foundation of our work. For multilingual settings, Sign2(LID+Text)~\citep{tan-etal-2025-multilingual} combines token-level sign language identification with a CTC objective to generate spoken text.

\subsection{Implementation Details}

\paragraph{$\bullet$ Feature Extraction.}
We begin by processing each sign language video $x = \{f_1, f_2, \dots, f_T\}$ as a sequence of $T$ RGB frames. From each frame, we extract:
\begin{itemize}
    \item \textbf{Spatial Features ($s_t$):} Using a Vision Transformer (ViT \citep{vitdosovitskiy2020image}) pretrained on ImageNet.
    \item \textbf{Motion Features ($m_t$):} Using VideoMAE \citep{tong2022videomae}.
\end{itemize}
These features are then fused via the visual fusion block $\mathcal{F}$ from SpaMo to yield a joint representation $h_t$ for each timestep.

\paragraph{$\bullet$ Training the visual block (LoRA).}
We train the visual block using LoRA with:
\begin{itemize}
  \item \textbf{LoRA:} \(r = 16,\ \alpha = 32\)
  \item \textbf{Batching:} batch size 4, gradient accumulation 2, on 8 GPUs 
  in parallel
  \item \textbf{Loss weights:}
  \begin{itemize}[leftmargin=1em]
    \item $\lambda_{\text{ce}}$ \(= 0.1\) (auxiliary soft translation signal)
    \item $\lambda_{\text{sem}}$ \(= 1.0\) (primary objective)
    \item $\lambda_{\cos}$ \(= 2.7\) (stabilizes angular alignment)
    \item $\lambda_{\text{nce}}$ \(= 0.0\)
    \item $\lambda_{\text{mse}}$ \(= 7000.0\) (strong magnitude regularizer)
  \end{itemize}
\end{itemize}

Because our model operates on embedding vectors with small magnitudes, the MSE loss can rapidly fall to $\sim 10^{-5}$ even when cosine similarity remains suboptimal. Empirically, we observed that \textbf{cosine and MSE only begin to correlate at $\sim 10^{-6}$}. Optimizing cosine alone often stalls, as MSE ceases to decrease, while optimizing MSE alone improves fidelity but does not guarantee angular alignment. To address this, we up-weight MSE to maintain shrinkage and retain a non-negligible cosine term to enforce directional consistency. We also experimented with InfoNCE, but under our effective batch size (with few hard negatives) it led to slower convergence and negligible improvements and we do not use it in our final experiments.

\paragraph{$\bullet$ Sentence embedding pooling.}
The original SONAR pools by running a shallow decoder: it feeds a special token (the EOS id in \texttt{M200M100}) as input and uses the encoder outputs as hidden states; the first decoder output is taken as the sentence embedding. During the Visual Block training, we adopt this approach with a shallow decoder initialized from the first three SONAR decoder layers and train it only for pooling. This supplies language context during pooling, while the incoming features themselves are language-agnostic (from another modality). Text generation is then conditioned on the target language.

\begin{table*}[ht]
 \small
\centering
\resizebox{\textwidth}{!}{
\begin{tabular}{lcccccc}
\toprule
\multirow{2}{*}{\textbf{Method}} & \multicolumn{3}{c}{\textbf{PHOENIX-2014T}} & \multicolumn{3}{c}{\textbf{CSL-Daily}} \\
\cmidrule(lr){2-4} \cmidrule(lr){5-7}
 & \textbf{BLEU} & \textbf{BLEURT} & \textbf{RG} & \textbf{BLEU} & \textbf{BLEURT} & \textbf{RG} \\
\midrule
\multicolumn{7}{l}{\textit{Monolingual}} \\
CSGCR \citep{zhao2021conditional} & 15.18 & -- & 38.85 & -- & -- & -- \\
GFSLT-VLP \citep{Zhou_2023_ICCV} & 21.44 & -- & 42.29 & 11.00 & -- & 36.44 \\
FLa-LLM \citep{chen-etal-2024-factorized} & 23.09 & -- & 45.27 & 14.20 & -- & 37.25 \\
Sign2GPT \citep{wong2024sign2gpt} & 22.52 & -- & 48.90 & 15.40 & -- & \textbf{42.36} \\
SignLLM \citep{gong2024signllm} & 23.40 & -- & 44.49 & 15.75 & -- & 39.91 \\
SEM-SLT \citep{hamidullah2024slt} & 24.10 & 0.481 & -- & -- & -- & -- \\
\midrule
\multicolumn{7}{l}{\textit{Multilingual}} \\
Sign2(LID+Text) \citep{tan-etal-2025-multilingual} & \textbf{24.23} & -- & \textbf{50.60} & 14.18 & -- & 40.00 \\
\textbf{SONAR-SLT (Ours)} & 22.01 & \textbf{0.545} & 41.44 & 16.23 & \textbf{0.561} & 42.29 \\
\bottomrule
\end{tabular}
}
\caption{Comparison of SONAR-SLT with other gloss-free models on PHOENIX-2014T and CSL-Daily (metrics: BLEU, BLEURT, ROUGE (RG)). Unreported metrics are left blank; SONAR-SLT sets the best reported BLEURT on PHOENIX-2014T and remains strongly competitive with several LLM-based baselines on both datasets.}
\label{tab:ph_csl_results_grouped}
\end{table*}

\paragraph{$\bullet$ Visual representation.}  
We adopt the best-performing visual representation strategies reported in prior work, noting that optimal choices vary across datasets. To ensure comparability in our multi–sign language experiments, we restrict evaluation to datasets with similar video settings and select the strongest corresponding model. The SpaMo Visual Block performs best with global, high-quality cues e.g., high-resolution videos with a moderate signer--camera distance (CSL-Daily), or lower-resolution videos where the signer is close and centered (PHOENIX-2014T). Consequently, we conduct multilingual experiments on CSL-Daily and PHOENIX-2014T.

\sisetup{
  table-format=2.2,
  detect-weight=true,
  detect-inline-weight=math
}

\sisetup{
  table-format=2.3,
  detect-weight=true,
  detect-inline-weight=math
}

\paragraph{$\bullet$ Training the translation model with visual features.}
We train the end-to-end translation system (with the Visual Block or the fused spatial+motion features) using the \textbf{same LoRA configuration} as above.

\begin{itemize}
  \item \textbf{Batching \& schedule:} batch size 8 on a single GPU
    \begin{itemize}[leftmargin=1em]
      \item \textbf{CSL-Daily:} cosine learning-rate schedule with a peak LR of \(3\times 10^{-4}\)
      \item \textbf{PHOENIX-2014-T (monolingual):} constant LR (we found it more stable)
    \end{itemize}

  \item \textbf{Text augmentation.} To expand the datasets using NLLB \citep{NLLBTeam2024Nature}, we machine-translate the target texts into three high-resource languages (English, French and Spanish) using the \texttt{facebook/nllb-200-distilled-600M} model. 

  \item \textbf{Video augmentation.}
Coupled with the target-language augmentation, we also perturb the input videos so that each training instance 
is presented with both linguistic and visual variability. 
At each iteration, one augmented variant is sampled. 
In this work we restrict ourselves to:  
\begin{itemize}[leftmargin=1em]
  \item \texttt{frame\_mask\_ratio} \(= 0.2\)  
  \item \texttt{frame\_dropout\_prob} \(= 0.2\)  
  \item \texttt{add\_noise\_std} \(= 0.04\)  
  \item \texttt{shuffle\_window} \(= 3\)  
\end{itemize}
\end{itemize}

\section{Results and Analysis}
\subsection{Comparative Analysis} 
We compare our approach with other gloss-free methods on both PHOENIX-2014T and CSL-Daily datasets in Table \ref{tab:ph_csl_results_grouped}.
Our method shows a clear advantage on the semantics-oriented BLEURT metric. It reaches a \textbf{BLEURT of 0.545}, outperforming the sentence-based supervision model using text-only sentence embedding (SEM-SLT). BLEURT uses a BERT-based scorer and is designed to capture meaning and fluency, unlike BLEU and ROUGE, which primarily measure $n$-gram overlap. Moreover, our model outperforms previous monolingual and multilingual systems on CSL-Daily in terms of BLEU and achieves comparable results on PHOENIX-2014T. 

\paragraph{$\bullet$ Observed gaps.}
We observe a decrease in BLEU compared to the SEM-SLT system, which is expected since our model is not fine-tuned on sign-language text. Our language-agnostic, sentence embedding-based supervision preserves semantics without requiring fine-tuning on specific dataset: it goes beyond surface $n$-gram matching to produce translations that are contextually accurate, grammatically correct, and cross-lingually robust. Part of the remaining gap stems from dataset capture conditions. Our feature extractor \citep{hwang-etal-2025-spamo} is tuned for global cues and can be less accurate in cases where fine-grained articulations, such as facial expressions and finger movements, are critical. Recent top systems address this with keypoint-based representations and extensive preprocessing \citep{shubert}, which help preserve these fine-grained details.


\begin{table}[t]
\centering
\setlength{\tabcolsep}{8pt}

\begin{tabular}{llccc}
\toprule
\textbf{Resource} & \textbf{Language} & \textbf{BLEU} \\
\midrule
\multirow{3}{*}{High}
  & Spanish (es)   & 22.3  \\
  & French (fr)    & 22.6 \\
  & English (en)   & 21.6 \\
\addlinespace[2pt]
\multirow{3}{*}{Low}
  & Turkish (tr)  & 13.1 \\
  & Malagasy (mg) & 11.8 \\
  & Persian (fa)  & ~8.7 \\
\bottomrule
\end{tabular}
\caption{SONAR-SLT performance across target languages in both high- and low-resource settings on PHOENIX-2014T, reported using BLEU scores.}
\label{tab:low_high_result}
\end{table}

\begin{table*}[ht]
\centering
\small
\setlength{\tabcolsep}{6pt}
\renewcommand{\arraystretch}{1.15}

\begin{tabular}{
    c
    l
    S
    S
    S
    S
    S
    S
}
\toprule
& & \multicolumn{3}{c}{\textbf{PHOENIX-2014T}} & \multicolumn{3}{c}{\textbf{CSL-Daily}} \\
\cmidrule(lr){3-5} \cmidrule(lr){6-8}
\textbf{Type} & \textbf{Variant} 
 & {\textbf{BLEURT}} & {\textbf{BLEU}} & {\textbf{RG}}
 & {\textbf{BLEURT}} & {\textbf{BLEU}} & {\textbf{RG}} \\
\midrule
\multirow{3}{*}{\rotatebox{90}{Multi}}
  & VB pretrained & \bfseries 0.523 &  21.52 & 41.10 
                   & \bfseries 0.561 & \bfseries 16.23 & \bfseries 42.29 \\
  & VB scratch    & 0.508 & 21.38 & \bfseries 42.03 
                   & 0.472 & 14.68 & 42.12 \\
  & VB frozen     & 0.516 & \bfseries 21.56 & 41.39 
                   & 0.549 &  16.06 & 41.95 \\
\midrule
\multirow{3}{*}{\rotatebox{90}{Mono}}
  & VB pretrained & \bfseries 0.545 & \bfseries 22.01 & 40.52 
                   & \bfseries 0.558 & \bfseries 16.07 & \bfseries 42.13 \\
  & VB scratch    & 0.490 & 19.79 & 39.95 
                   & 0.447 & 14.14 & 40.59 \\
  & VB frozen     & 0.520 & 21.56 & \bfseries 41.44 
                   & 0.529 & 15.70 & 41.79 \\
\bottomrule
\end{tabular}
\caption{SONAR-SLT results for the Visual Block (VB) variants under Multilingual and Monolingual settings on PHOENIX-2014T and CSL-Daily. Metrics include BLEURT, BLEU, and ROUGE (RG); best scores per dataset/metric are in bold.}
\label{tab:vb-main}
\end{table*}

\paragraph{$\bullet$ Multilingual and multi–sign language.}
We evaluate target-side augmentation, where language translations are included in training. Results for both low- and high-resource languages on PHOENIX-2014T are presented in Table~\ref{tab:low_high_result}. In our experiments, we augmented the target set in training with three high-resource languages—French, Spanish, and English—while the model was evaluated on other unseen languages. Using this augmented target set yields a modest improvement over training with a single target language. However, we observe a gap in performance between high- and low-resource languages, which primarily stems from lower reference translation quality in the low-resource languages.\footnote{Reference translations were obtained using \texttt{facebook/nllb-200-distilled-600M} model.} The narrow domain of PHOENIX-2014T can also introduce dataset-specific idiosyncrasies, complicating fair comparisons.

Table \ref{tab:vb-main} shows that pre-training on concatenated multi-sign corpora followed by monolingual fine-tuning proves most effective. In contrast, joint multi-sign fine-tuning risks resembling another full training run without yielding substantial gains. In our experiments, we first pre-train on the combined data and then fine-tune monolingually, consistent with~\citep{hamidullah2024slt}; post-fine-tuning performance remains largely unchanged (see Table~\ref{tab:vb-main}, mono vs. multilingual setup). Differences in dataset capture conditions still matter—for example, methods that rely solely on global visual features can underperform when fine-grained articulations, such as hand or facial details, are crucial. Pipelines that integrate keypoints with extensive preprocessing~\citep{shubert} help mitigate such losses and achieve stronger results.

\subsection{Multitask Learning Effect on the Visual Block}
The effect of sentence-embedding supervision is strongest when the Visual Block is still learning feature representations. 
Once the block has converged—or is pretrained—the additional impact of cosine or MSE objectives diminishes. 
This occurs because cross-entropy loss often remains relatively high (above $2$--$3$), 
while MSE rapidly falls to $\mathcal{O}(10^{-5})$ and cosine similarity saturates around $\sim 0.3$.  

In contrast, introducing the auto-encoding loss provides a second cross-entropy signal, 
which exerts a stronger influence on the Visual Block. 
Here, intermediate supervision continues to be beneficial, and the auto-encoding objective itself 
accelerates convergence. 
We consistently observed this effect in CSL-Daily and in the augmented translation setup on PHOENIX-2014T.

\subsection{Qualitative and Semantic Error Analysis}
\paragraph{$\bullet$ Qualitative analysis.}
Table~\ref{tab:de_fr_examples} shows examples of two contrasting outcomes: cases where the model accurately captures the intended meaning and cases where it fails. 
When contextual understanding is incomplete, the decoder frequently compensates by generating fluent continuations via next-token prediction. 
This behavior is characteristic of SLT systems that rely on pretrained language models as decoders: they can mask weaknesses in semantic grounding by producing outputs that are coherent but only partially faithful to the source. 
As a result, improvements in BLEU may reflect the decoder’s ability to recover plausible sentences rather than true gains in sign-to-text comprehension. 
Therefore, exact sequence matching metrics such as BLEU are insufficient and in some cases misleading for evaluating translation quality in SLT. 

\textbf{Language-specific tendencies.} We deep into the analysis of two languages: German, a language trained with original data and French, a language trained via machine translation augmentation.
\begin{itemize}
    \item \textbf{German:} Errors often arise from compound nouns, flexible word order, and embedded clauses, leading to partial omissions, 
    attribute reordering, or unnatural compounds. When alignment is uncertain, the model may insert generic stock phrases or repetitions.  
    \item \textbf{French:} Our analysis shows more frequent noun substitutions, agreement mismatches, and text modality shifts (e.g., hedging with \textit{``sont possibles''}). 
    Register differences from determiners or prepositions are also common. Incorrect date and numeric substitutions occur more frequently than in German, 
    likely due to segmentation differences in temporal expressions.  
\end{itemize}

\paragraph{$\bullet$ Semantic analysis.} Surface-form scores vs.\ meaning preservation.
We observe a systematic mismatch between surface-form metrics (e.g., BLEU) and semantic adequacy (BLEURT) across both German and French. 
Outputs with only moderate $n$-gram overlap can still be semantically faithful, while some high-scoring predictions contain factual errors.  

\textbf{Semantically near correct and correct paraphrases (German).}  
As illustrated by the green-highlighted examples in Table~\ref{tab:de_fr_examples}, incorrect lexical or numeric substitutions leave most of the remaining meaning intact 
(e.g., date shifts: \textit{``Sonntag, den neunzehnten Dezember''} $\rightarrow$ \textit{``Sonntag, den siebzehnten August''}; 
temperature adjustments: \textit{``sechs Grad an den Alpen''} $\rightarrow$ \textit{``neun Grad am Alpenrand''}). 
We also observe benign stylistic reformulations (\textit{``es gelten entsprechende Warnungen''} $\rightarrow$ \textit{``es bestehen Unwetterwarnungen''}) 
and word-order changes without semantic effect (\textit{``aus Südwest bis West''} $\rightarrow$ \textit{``aus West bis Südost''}).  

\textbf{Semantically near correct and correct paraphrases (French).}  
Similarly, the first green-highlighted examples show structural or modality shifts that preserve much of the remaining meaning, such as date substitutions 
(\textit{``vingt-huit août''} $\rightarrow$ \textit{``vingt-cinq novembre''}), 
modality changes (\textit{``Des rafales orageuses de l’ouest''} $\rightarrow$ \textit{``Des rafales orageuses sont possibles''}), 
or expanded phrasing (\textit{``Également orages sur la mer du Nord''} $\rightarrow$ \textit{``Il y a également des orages sur la mer du Nord''}).  

\textbf{Semantically incorrect outputs, true errors (French and German).} 
The red-highlighted rows in Table~\ref{tab:de_fr_examples} illustrate errors such as topic drift (predicting wind instead of temperature), 
incorrect locations (\textit{``Höhenlagen Süddeutschlands''} $\rightarrow$ \textit{``Küsten''}; 
\textit{``sud-est''} $\rightarrow$ \textit{``nord''}), 
system inversions (\textit{``Hoch''} $\leftrightarrow$ \textit{``Tief''}; \textit{``haut''} $\leftrightarrow$ \textit{``profonde''}), 
hallucinated entities, or incorrect hazard categories 
(\textit{``Risque d’inondation''} $\rightarrow$ \textit{``Avertissements météorologiques violents''}).  

Overall, as in other machine translation tasks, $n$-gram metrics penalize near or even fully legitimate paraphrases and sometimes fail to capture serious factual errors. 
Robust SLT evaluation requires \emph{semantic} metrics that explicitly reward meaning preservation while penalizing distortions or hallucinations.

\begin{table}[!t]
\centering
\resizebox{\linewidth}{!}{%
\begin{tabular}{p{10cm}}
\toprule
\textbf{German (DGS weather domain)} \\
\midrule

\rowcolor{green!10} \textbf{Ref (DE):} Sonntag, den neunzehnten Dezember. \\
\rowcolor{green!10} \textit{EN: Sunday, the nineteenth of December.} \\
\rowcolor{green!10} \textbf{Pred (DE):} Sonntag, den siebzehnten August. \\
\rowcolor{green!10} \textit{EN: Sunday, the seventeenth of August.} \\
\midrule

\rowcolor{green!10} \textbf{Ref (DE):} sechs Grad an den Alpen. \\
\rowcolor{green!10} \textit{EN: Six degrees in the Alps.} \\
\rowcolor{green!10} \textbf{Pred (DE):} neun Grad am Alpenrand. \\
\rowcolor{green!10} \textit{EN: Nine degrees on the edge of the Alps.} \\
\midrule

\rowcolor{red!10} \textbf{Ref (DE):} Höhenlagen Süddeutschlands. \\
\rowcolor{red!10} \textit{EN: High-altitude areas of southern Germany.} \\
\rowcolor{red!10} \textbf{Pred (DE):} Küsten. \\
\rowcolor{red!10} \textit{EN: Coasts.} \\
\bottomrule
\end{tabular}%
}

\vspace{0.6em}

\resizebox{\linewidth}{!}{%
\begin{tabular}{p{10cm}}
\textbf{French (weather domain)} \\
\midrule

\rowcolor{green!10} \textbf{Ref (FR):} vingt-huit août. \\
\rowcolor{green!10} \textit{EN: Twenty-eighth of August.} \\
\rowcolor{green!10} \textbf{Pred (FR):} vingt-cinq novembre. \\
\rowcolor{green!10} \textit{EN: Twenty-fifth of November.} \\
\midrule

\rowcolor{green!10} \textbf{Ref (FR):} Des rafales orageuses de l’ouest. \\
\rowcolor{green!10} \textit{EN: Stormy gusts from the west.} \\
\rowcolor{green!10} \textbf{Pred (FR):} Des rafales orageuses sont possibles. \\
\rowcolor{green!10} \textit{EN: Stormy gusts are possible.} \\
\midrule

\rowcolor{red!10} \textbf{Ref (FR):} Risque d’inondation. \\
\rowcolor{red!10} \textit{EN: Risk of flooding.} \\
\rowcolor{red!10} \textbf{Pred (FR):} Avertissements météorologiques violents. \\
\rowcolor{red!10} \textit{EN: Severe weather warnings.} \\
\bottomrule
\end{tabular}%
}
\caption{German and French examples —two semantically near correct paraphrases (green) and one semantically incorrect output (red), with English translations.}
\label{tab:de_fr_examples}
\end{table}

\paragraph{Implications.}
Evaluation and model development for multilingual SLT should be language-aware. In practice, one should combine semantics-focused metrics with targeted, language-specific checks (e.g., temporals and agreement in French; word order and compounding in German) to obtain fair comparisons and actionable diagnostics.
\section{Conclusion}
We presented a scalable SLT framework that breaks the traditional close dependency between sign and spoken languages in training data and system development. 
By aligning sign language videos with multilingual, multimodal sentence embeddings from \textsc{SONAR}, our approach yields a 
language-agnostic semantic representation that generalizes across both sign languages and spoken targets. 
This reduces reliance on language-model priors and prioritizes visual grounding and SLT-specific grammar over surface-level text patterns.  

Experiments show that language-agnostic supervision enables robust translation even under sign--target mismatches. 
Multilingual text augmentations, combined with visual augmentation, improves performance on PHOENIX-2014T despite 
limited data. Ablations further confirm the advantages of this approach in preserving semantic adequacy.  

Current evaluation practices often emphasize surface overlap rather than meaning. 
Future work should develop metrics aligned with semantic similarity and extend supervision to 
low-resource sign languages and continuous signing in the wild.  

\section*{Limitations}
Our main limitation lies in the visual feature extractor rather than the model architecture itself. We used a pre-existing visual block to avoid evaluation bias, which restricted us to datasets with compatible video settings (CSL-Daily and PHOENIX-2014T) and excluded larger corpora such as How2Sign or YouTube-ASL. As a result, our approach focuses on preserving semantics rather than maximizing exact sentence matches.

Machine translation for data augmentation might induce unintended cultural mistakes that go beyond literal translation. The evaluation on non-human translated datasets also limits the strength of the conclusions for low-resourced languages.

\section*{Acknowledgments}
This work was partially funded by the German ministry for education and research (BMBF) through projects BIGEKO (grant number 16SV9093) and TRAILS (grant number 01IW24005).

\bibliography{custom}

\clearpage

\appendix

\section{Appendix}
\label{sec:appendix}

\subsection{Detailed Architecture}
\label{app:arch}
\begin{figure}[h]
    \centering
    \includegraphics[width=3.1\linewidth,angle=90]{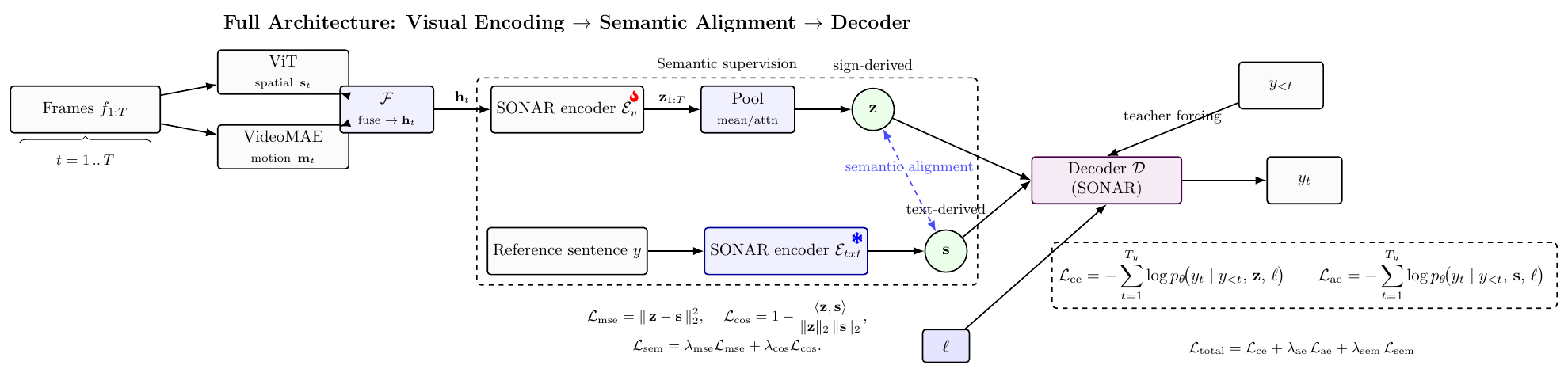}
    \caption{Detailed architecture without the contrastive term (NCE loss).}
    \label{fig:detailed_architecture}
\end{figure}

Figure~\ref{fig:detailed_architecture} shows the details of our system architecture as explained in Section~\ref{sec:method}.

\clearpage
\newpage

\subsection{Porting SONAR from NLLB Fairseq to Huggingface.}
SONAR is officially supported in fairseq, but only its text encoder is available on Hugging Face. To enable full conditional generation, we ported both the encoder and decoder weights from the original SONAR checkpoints into \texttt{M200M100}, extending the earlier encoder-only port provided by the NLLB team. In particular, we transferred the decoder weights directly from fairseq, validated their functionality, and released the complete model for public use. The resulting \texttt{M200M100ForConditionalGeneration} can now be loaded end-to-end and fine-tuned directly.

\newpage
\subsection{Additional Qualitative Results}
Additional translation examples for CSL-Daily and PHOENIX-2014T are provided in Tables~\ref{tab:csl_examples} and~\ref{tab:phoenix_examples}, respectively.

\begin{table}[!h]
\centering
\resizebox{\linewidth}{!}{%
\begin{tabular}{p{9cm}}
\hline
\textbf{Text} \\
\hline
\addlinespace[3pt] 
\rowcolor{green!10} \textbf{Ref (ZH):} \begin{CJK*}{UTF8}{gbsn}我们下午三点见面。\end{CJK*} \\
\rowcolor{green!10} \textit{EN: We will meet at three in the afternoon.} \\
\rowcolor{green!10} \textbf{Pred (ZH):} \begin{CJK*}{UTF8}{gbsn}我们三点钟下午见。\end{CJK*} \\
\rowcolor{green!10} \textit{EN: We meet at three o’clock in the afternoon.} \\
\cline{1-1}
\addlinespace[3pt] 
\rowcolor{green!10} \textbf{Ref (ZH):} \begin{CJK*}{UTF8}{gbsn}我早上吃面包和牛奶。\end{CJK*} \\
\rowcolor{green!10} \textit{EN: I eat bread and milk in the morning.} \\
\rowcolor{green!10} \textbf{Pred (ZH):} \begin{CJK*}{UTF8}{gbsn}我早上吃了牛奶和面包。\end{CJK*} \\
\rowcolor{green!10} \textit{EN: I had milk and bread in the morning.} \\
\cline{1-1}
\addlinespace[3pt] 
\rowcolor{red!10} \textbf{Ref (ZH):} \begin{CJK*}{UTF8}{gbsn}我们乘坐飞机去旅游，今天在酒店住宿。\end{CJK*} \\
\rowcolor{red!10} \textit{EN: We took a plane to travel, and are staying in a hotel today.} \\
\rowcolor{red!10} \textbf{Pred (ZH):} \begin{CJK*}{UTF8}{gbsn}我们飞机去上海，今天喝酒睡觉。\end{CJK*} \\
\rowcolor{red!10} \textit{EN: We took a plane to Shanghai, today we drink alcohol and sleep.} \\
\hline

\end{tabular}
}
\caption{CSL-Daily examples ---good translations (green) and one bad translation (red), showing reference and prediction in Chinese, with English translations for clarity.}
\label{tab:csl_examples}
\end{table}

\begin{table}[!h]
\centering
\resizebox{\linewidth}{!}{%
\begin{tabular}{p{10cm}}
\toprule
\textbf{Text} \\
\midrule

\rowcolor{green!10} \textbf{Ref (DE):} ich wünsche ihnen noch einen schönen abend. \\
\rowcolor{green!10} \textit{EN: I wish you a pleasant evening.} \\
\rowcolor{green!10} \textbf{Pred (DE):} und jetzt wünsche ich ihnen noch einen schönen abend. \\
\rowcolor{green!10} \textit{EN: And now, I wish you a pleasant evening.} \\
\rowcolor{green!10} \textbf{Pred (FR):} Et maintenant, je vous souhaite une bonne soirée. \\
\rowcolor{green!10} \textit{EN: And now, I wish you a good evening.} \\
\midrule

\rowcolor{green!10} \textbf{Ref (DE):} der wind aus süd bis west weht schwach bis mäßig. \\
\rowcolor{green!10} \textit{EN: The wind from the south to west blows weakly to moderately.} \\
\rowcolor{green!10} \textbf{Pred (DE):} der wind weht meist schwach aus süd bis west. \\
\rowcolor{green!10} \textit{EN: The wind generally blows weakly from south to west.} \\
\rowcolor{green!10} \textbf{Pred (FR):} Le vent souffle généralement faiblement du sud-ouest. \\
\rowcolor{green!10} \textit{EN: The wind generally blows weakly from the southwest.} \\
\midrule

\rowcolor{red!10} \textbf{Ref (DE):} in deutschland gibt es nur schwache luftdruckunterschiede. \\
\rowcolor{red!10} \textit{EN: In Germany, there are only slight air pressure differences.} \\
\rowcolor{red!10} \textbf{Pred (DE):} im nordosten deutschlands sorgt das hoch für wenig unbeständiges wetter. \\
\rowcolor{red!10} \textit{EN: In northeastern Germany, the high pressure system causes little unsettled weather.} \\
\rowcolor{red!10} \textbf{Pred (FR):} Dans certaines régions de l'Allemagne, la pression atmosphérique élevée n'est toujours pas atteinte. \\
\rowcolor{red!10} \textit{EN: In some regions of Germany, high atmospheric pressure has still not been reached.} \\
\bottomrule
\end{tabular}%
}
\caption{PHOENIX-2014T examples ---two good translations (green) and one bad translation (red), showing reference (German), predictions (German and French), and English translations.}
\label{tab:phoenix_examples}
\end{table}

\end{document}